\documentclass[twoside,11pt]{article}

\usepackage{jmlr2e}
\usepackage{graphicx}
\usepackage{amsmath}
\usepackage{natbib}
\usepackage{amssymb}
\usepackage{algorithm,algorithmic}
\usepackage{epstopdf}
\usepackage{footnote}

\ShortHeadings{A Library for Scalable Second Order Learning Algorithms}{Vinod, Anuj and Kalpana}
\firstpageno{1}

\begin{document}

\title{LIBS2ML: A Library for Scalable Second Order\\ Machine Learning Algorithms}
\author{\name Vinod Kumar Chauhan \email vkumar@pu.ac.in\\
	\addr Homepage: \url{https://sites.google.com/site/jmdvinodjmd}
	\AND
	\name Anuj Sharma \email anujs@pu.ac.in\\
	\addr Department of Computer Science \& Applications,\\Panjab University Chandigarh, India\\
	Homepage: \url{https://sites.google.com/view/anujsharma}
	\AND
	\name Kalpana Dahiya \email kalpanas@pu.ac.in\\
	\addr University Institute of Engineering \& Technology,\\Panjab University Chandigarh, India
}

\editor{...}

\maketitle

\begin{abstract}
LIBS2ML\footnotemark is a library based on scalable second order learning algorithms for solving large-scale problems, i.e., big data problems in machine learning. LIBS2ML has been developed using MEX files, i.e., C++ with MATLAB/Octave interface to take the advantage of both the worlds, i.e., faster learning using C++ and easy I/O using MATLAB. Most of the available libraries are either in MATLAB/Python/R which are very slow and not suitable for large-scale learning, or are in C/C++ which does not have easy ways to take input and display results. So LIBS2ML is completely unique due to its focus on the scalable second order methods, the hot research topic, and being based on MEX files. Thus it provides researchers a comprehensive environment to evaluate their ideas and it also provides machine learning practitioners an effective tool to deal with the large-scale learning problems. LIBS2ML is an open-source, highly efficient, extensible, scalable, readable, portable and easy to use library. The library can be downloaded from the URL: \url{https://github.com/jmdvinodjmd/LIBS2ML}.
\end{abstract}


\section{Introduction}
\label{sec_intro}
Nowadays big data is one of the major challenge in machine learning~\citep{Chauhan2018Review} and stochastic optimization techniques have been quite effective to tackle the challenge~\citep{Chauhan2019SAAGs34}. After the success of stochastic first order methods \citep{Chauhan2017Saag,Chauhan2018SS_AI}, the focus of learning algorithms have shifted towards the stochastic second order methods due to their faster convergence rates and availability of computing resources to deal with their high computational complexities~\citep{Chauhan2018STRON}.\\
\noindent In this library, we have solved the commonly used empirical risk minimization problem in machine learning, as given below:
\begin{equation}
\label{eq_erm}
\min_{w} F(w) = \dfrac{1}{n} \sum_{i=1}^{n} f(w; x_i, y_i) = \dfrac{1}{n} \sum_{i=1}^{n} f_i(w),
\end{equation}
where $w \in \mathbb{R}^d$ is a parameter vector, $d$ is the number of features, $f_i(w): \mathbb{R}^d \rightarrow \mathbb{R}$ is a loss function and $\left\lbrace \left(x_i, y_i \right) \right\rbrace_{i=1}^{n}$ is the training set with $n$ data points.\\
\noindent LIBS2ML \footnotetext{LIBS2ML is continuously extended by adding more problems and methods.} has been developed due to great scope of scalable second order methods and make the task of researchers and practitioners easy. MATLAB/Python/R based libraries are easy to use, easy to code and explore new ideas but these are slow and so these are not suitable to deal with large-scale problems, e.g. SGDLibrary~\citep{Kasai2018}. On the other hand, C/C++ based libraries are fast and good to solve large-scale problems but they are difficult to code and lack functionality to show convergence of algorithms, e.g. LIBLINEAR~\citep{LIBLINEAR}. LIBS2ML combines the best of these two programming worlds, where all learning algorithms are implemented using C++ and MATLAB/Octave is used to take input and display results.

\section{Scalable Second Order Machine Learning}
\label{sec_main}
\subsection{Overview}
\label{subsec_overview}
In this library only major second order methods are considered which are suitable for solving large-scale learning problems. Currently, we have considered linear classification problem but more problems, like linear regression are being added to it. The second order algorithms can be divided mainly into two categories: Quasi-Newton methods~\citep{Byrd2011} and Inexact Newton methods~\citep{Chauhan2018STRON}. Both of which try to solve the issue of large complexities caused by the Hessian matrix by approximating the Hessian inverse and by solving the linear system inexactly.
Due to space limitations, it is not feasible to list all the learning algorithms supported by the library so please look at the library web page to see the complete list of algorithms. Moreover, the data formats have been kept same as that of LIBSVM \citep{LIBSVM} library for the convenience of users.

\subsection{The Software Package}
\label{subsec_soft_pack}
\subsubsection{Architecture and Documentation}
\label{subsec_doc_design}
\begin{figure}[htb]
	\centering
	\includegraphics[width=1\linewidth]{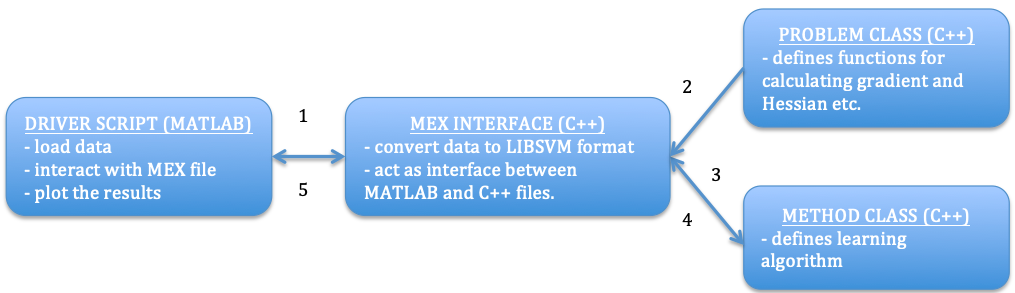}
	\caption{Modular design of LIBS2ML facilitates readability and easy extensibility.}
	\label{fig_design}
\end{figure}

LIBS2ML has been designed using a modular approach for flexibility and extensibility. The library separates problem, method, I/O and auxiliary code into different classes and folders. Each problem like, logistic regression, is contained in a separate C++ file which defines gradients, Hessian and related calculations. Similarly, each learning method is defined in a separate C++ file. Fig.~\ref{fig_design} depicts the high level architecture of LIBS2ML, which shows four components: First is MATLAB/Octave driver script which performs I/O, second is a MEX file which acts as an interface between MATLAB and C++, third and fourth components are problem and method classes, which are developed in C++.
The design of the library provides following features:\\
\noindent \textbf{Highly Efficient:} All the learning algorithms are developed in C++.\\
\noindent \textbf{Modular Design:} The library separates problems, methods and I/O operations.\\
\noindent \textbf{Extensibility:} New problem or method can be added by adding single new file.\\
\noindent \textbf{Portability:} The library compiles and runs on MATLAB/Octave so it is portable and can work on any platform (Windows, Mac OS and Linux etc.).\\
\noindent \textbf{Easy:} The library is easy to install, use and understand, and has no dependency on other libraries.\\
\noindent The library is thoroughly documented. It contains a README file which provides detailed documentation for using the library with following sections: `INSTALLATION' contains instructions for installing the library on different platforms, `USAGE gives an example to use the library, `EXTENSION' section contains instructions for adding new features, like new problems and solvers, and `MORE INFORMATION' contains more related information.

\subsubsection{Practical Usage}
\label{subsec_usage}
\begin{figure}[htb]
	\centering
	\includegraphics[width=1\linewidth]{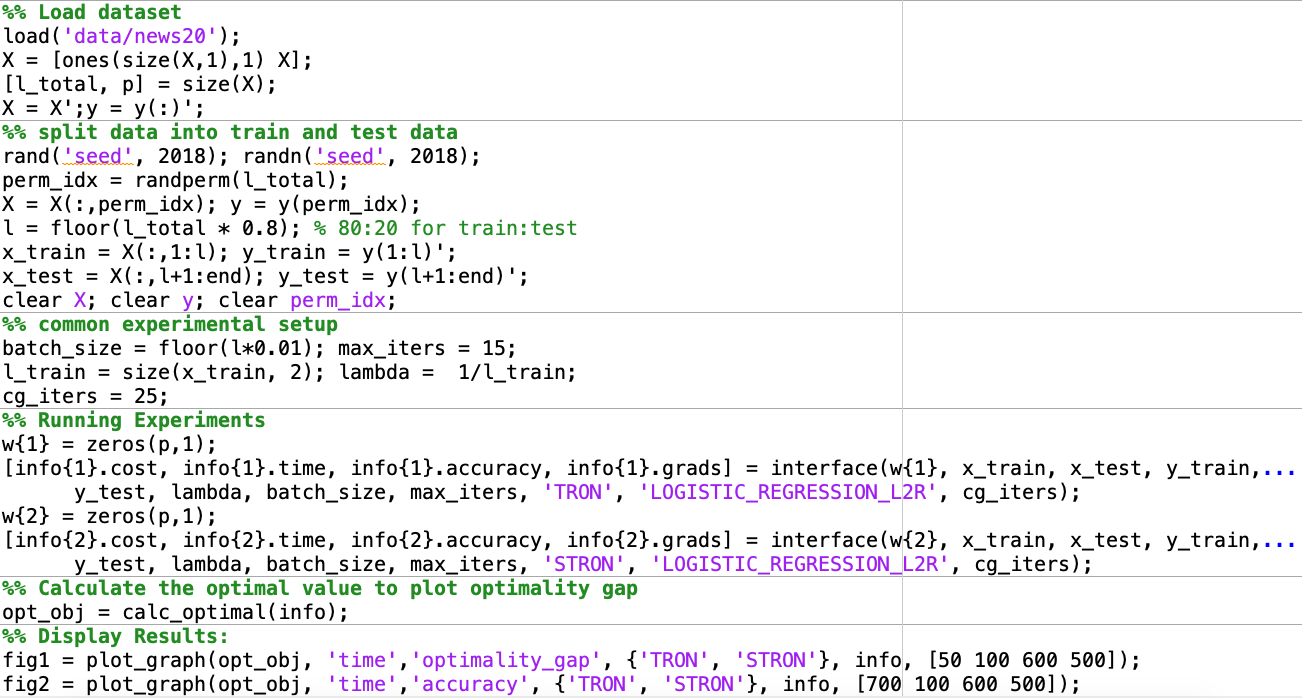}
	\caption{MATLAB driver script example.}
	\label{fig_demo}
\end{figure}

\begin{figure}[htb]
	\centering
%
	
	{\includegraphics[width=.485\linewidth]{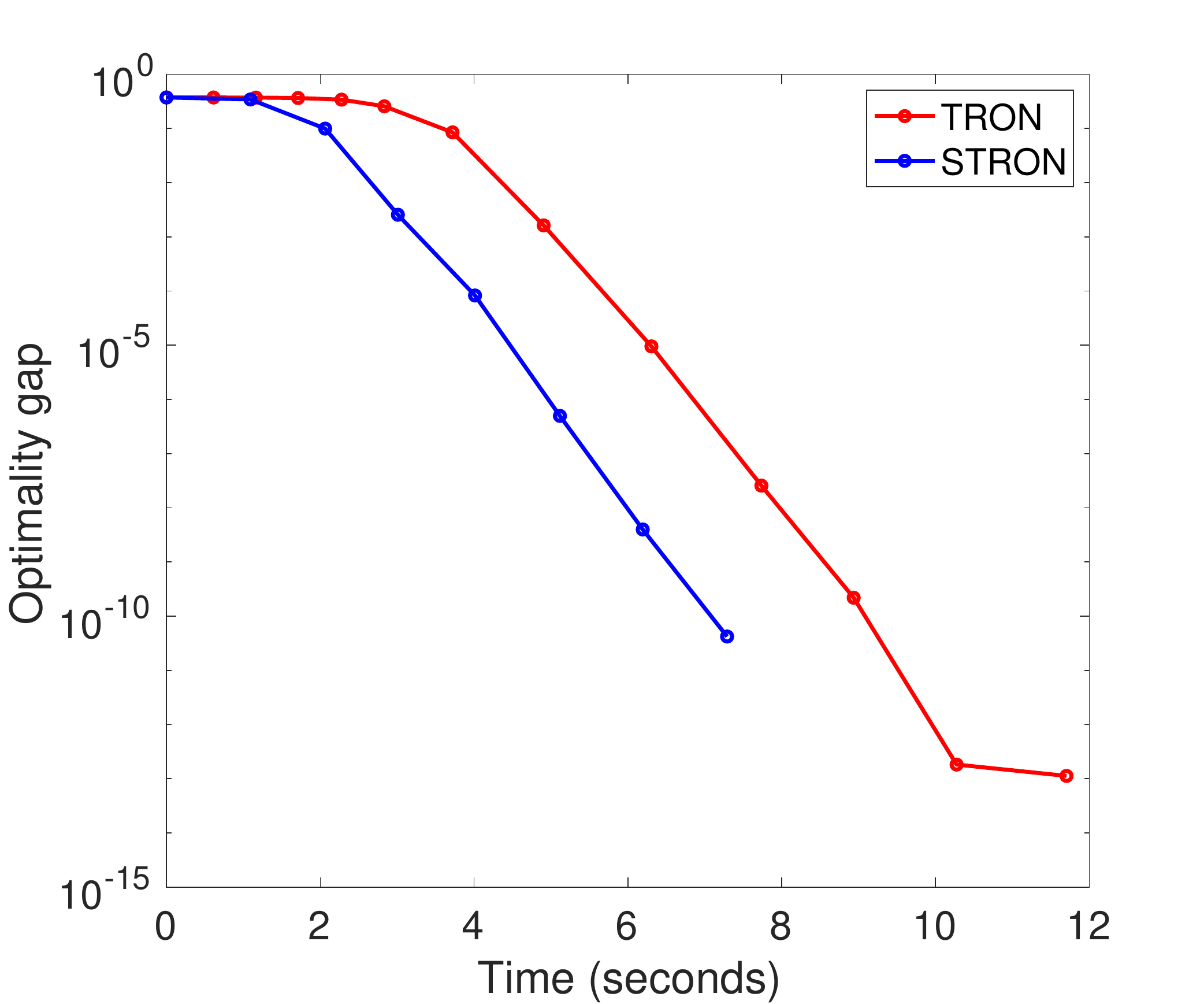}}
	{\includegraphics[width=.485\linewidth]{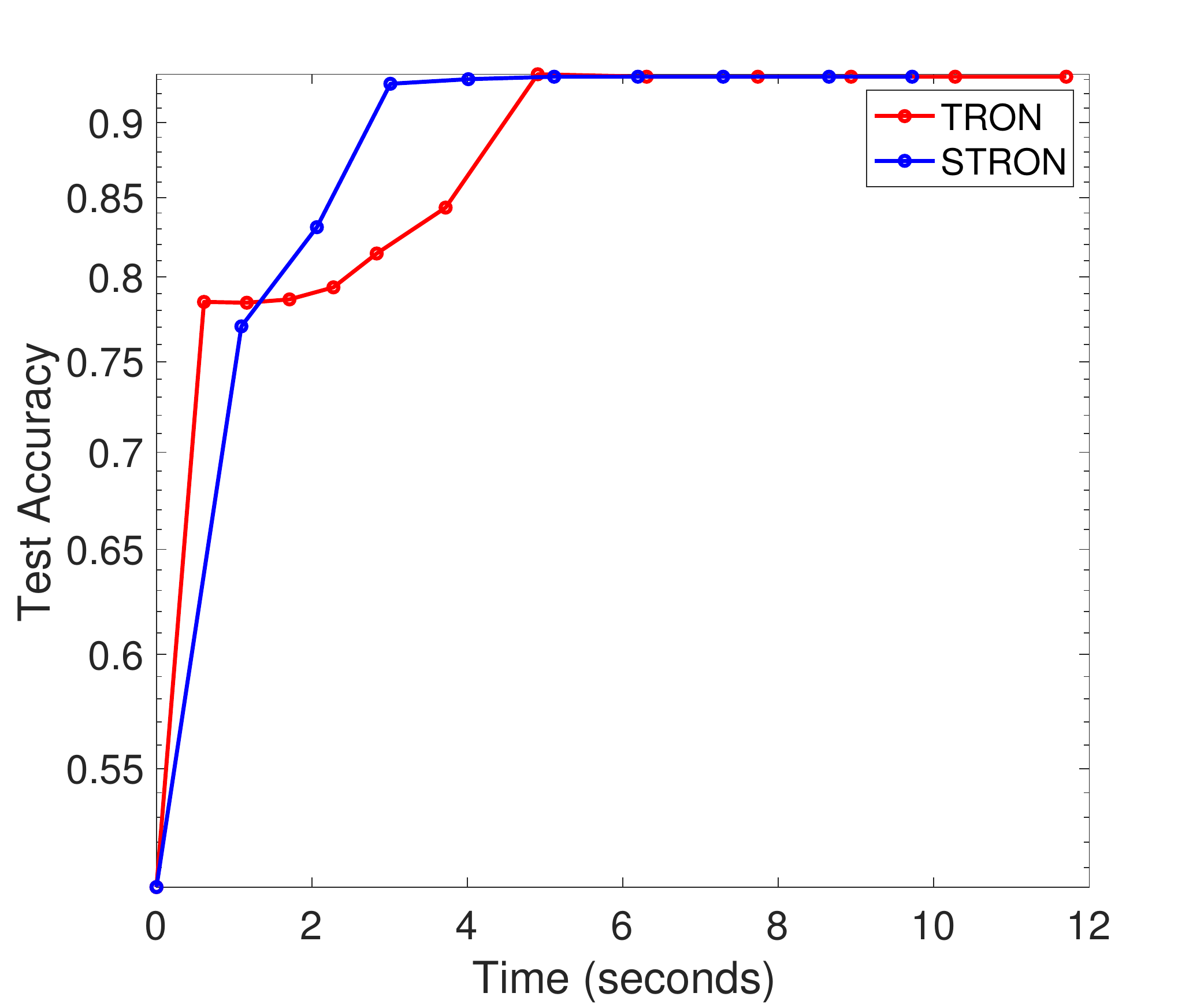}}
	
	\caption{TRON and STRON are compared on optimality and test accuracy against the training time.}
	\label{fig_comparison}
\end{figure}

MATLAB/Octave with a compatible C/C++ compiler are the prerequisites for LIBS2ML. Moreover, we need to develop a driver script in MATLAB/Octave to solve a problem using the library, where we load the desired dataset, call the MEX interface with information about the problem to be solved, learning method, data and other required parameters, and plot the results in the desired format. An example of practical usage of LIBS2ML has been given in Fig.~\ref{fig_demo}, which shows the driver script and the results are depicted with Fig.~\ref{fig_comparison}. This example solves the l2-regularized logistic regression problem using TRON~\citep{Hsia2018} and STRON~\citep{Chauhan2018STRON} methods with news20 dataset~\citep{Keerthi2005} and results are plotted as optimality and test accuracy against the training time (in seconds) of learning algorithms. The results can be plotted against number of epochs but that is not useful for large-scale learning problems because small and large-scale learning has different trade-offs, and time is a major factor for large-scale learning~\citep{Bottou2008}.

\section{Conclusion}
\label{sec_conclusion}
LIBS2ML is completely unique library due to its focus on scalable second order methods, the hot research topic, for solving large-scale learning problems, and using best of MATLAB and C++. It is modular, easily understandable, extensible, portable, highly efficient and specially designed to deal with the big data challenge in machine learning. It is helpful for beginners as well as advanced researchers to explore new ideas, and to the machine leaning practitioners.

\acks{First author is thankful to Ministry of Human Resource Development, Government of INDIA, to provide fellowship (University Grants Commission - Senior Research Fellowship) to pursue his PhD.}

\vskip 0.2in

\end{document}